\def\year{2020}\relax
\documentclass[letterpaper]{article} % DO NOT CHANGE THIS
\usepackage{aaai20}  % DO NOT CHANGE THIS
\usepackage{times}  % DO NOT CHANGE THIS
\usepackage{helvet} % DO NOT CHANGE THIS
\usepackage{courier}  % DO NOT CHANGE THIS
\usepackage[hyphens]{url}  % DO NOT CHANGE THIS
\usepackage{graphicx} % DO NOT CHANGE THIS
\usepackage{graphicx}  % DO NOT CHANGE THIS
\usepackage{lipsum}
\usepackage{multirow}
\usepackage[dvipsnames]{xcolor}
\usepackage{amsmath,bm}
\usepackage{subcaption}
\usepackage{nicefrac}
\usepackage{hyperref}
\hypersetup{
    colorlinks=true,
    linkcolor=blue,
    filecolor=magenta,      
    urlcolor=cyan,
}
\DeclareMathAlphabet\mathbfcal{OMS}{cmsy}{b}{n}

\title{Symmetrical Synthesis for Deep Metric Learning}
\author{Geonmo Gu\thanks{Authors contributed equally.}, Byungsoo Ko\footnotemark[1]\\
NAVER/LINE Vision\\ %If you have multiple authors and multiple affiliations
\{korgm403, kobiso62\}@gmail.com % email address must be in roman text type, not monospace or sans serif
}

\begin{document}

\maketitle

\begin{abstract}
Deep metric learning aims to learn embeddings that contain semantic similarity information among data points.
To learn better embeddings, methods to generate synthetic hard samples have been proposed.
Existing methods of synthetic hard sample generation are adopting autoencoders or generative adversarial networks, but this leads to more hyper-parameters, harder optimization, and slower training speed.
In this paper, we address these problems by proposing a novel method of synthetic hard sample generation called symmetrical synthesis.
Given two original feature points from the same class, the proposed method firstly generates synthetic points with each other as an axis of symmetry.
Secondly, it performs hard negative pair mining within the original and synthetic points to select a more informative negative pair for computing the metric learning loss.
Our proposed method is hyper-parameter free and plug-and-play for existing metric learning losses without network modification.
We demonstrate the superiority of our proposed method over existing methods for a variety of loss functions on clustering and image retrieval tasks.
Our implementations is publicly available.\footnote{\url{https://github.com/clovaai/symmetrical-synthesis}}
\end{abstract}

\section{Introduction}

\begin{figure}[t]
\centering
\includegraphics[width=0.9\columnwidth]{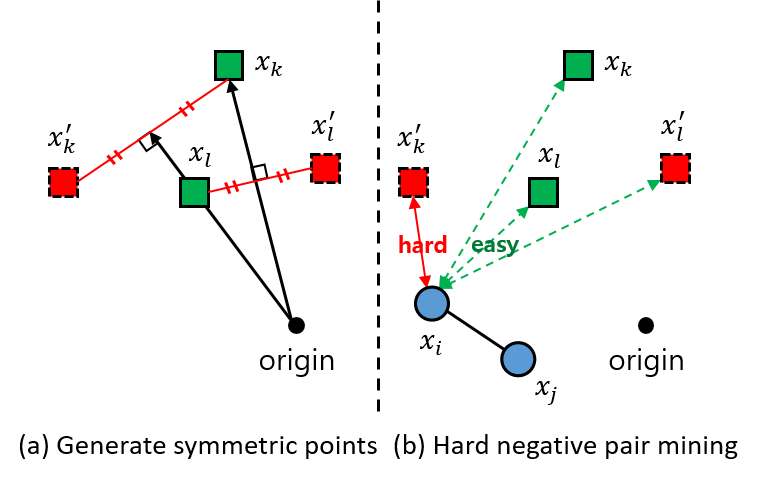}
\caption{Illustration of our proposed symmetrical synthesis with two steps.
First, given positive points ($\mathbf{x}_i$, $\mathbf{x}_j$) and negative points ($\mathbf{x}_k$, $\mathbf{x}_l$) in an embedding space, the negative points generate their synthetic points ($\mathbf{x}_k^{'}$, $\mathbf{x}_l^{'}$) with each other as an axis of symmetry.
% First, given negative feature points ($x_{n_1}$, $x_{n_2}$), each point generates synthetic points ($x_{n_1}^{'}$, $x_{n_2}^{'}$) which are symmetric with respect to vectors of the other points ($\vec{x_{n_1}}$, $\vec{x_{n_2}}$).
Secondly, it selects the hardest negative point within the four feature points: two original points and two synthetic points.
In the figure, $\mathbf{x}_k^{'}$ will be selected.
Rectangles and circles represent two different classes.
Green and blue points are original features while red points with dotted boundary are synthetic features.}
\label{fig:teaser}
\end{figure}

The objective of deep metric learning is to learn an embedding space where semantically similar images are embedded close together, and semantically dissimilar images are embedded far apart.
% Thanks to the great success of the deep neural network, a non-linear function that maps from a data space to an embedding space can be directly learned.
% The more closely the non-linear function reflects semantic information of the data, the higher performance we can expect.
% Deep metric learning has played an important role in a variety of computer vision tasks, such as image retrieval~\cite{gordo2017end,jun2019combination,ko2019benchmark}, re-identification~\cite{hermans2017defense,ahmed2015improved}, and clustering~\cite{HersheyCRW15}.
% Similarity or distance based metric learning loss is instrumental in deep metric learning.
Many recent deep metric learning approaches are built on similarity or distance between pairs of samples.
% Similarity or distance based metric learning loss has been introduced .
Contrastive loss~\cite{chopra2005learning} and triplet loss~\cite{weinberger2009distance} are both conventional losses that consider pair and triplet feature points, respectively.
Recent works~\cite{sohn2016improved,oh2016deep,wang2017deep} have modified the structures of loss functions to contain richer information by considering multiple feature points, and they have achieved competitive performances.
Along with the importance of the loss function, sampling strategy is also known to be essential for effective training.
Different sampling strategies can lead to drastically different performances for the same loss function.
This has motivated recent works to focus on sampling strategies, such as hard negative pair mining~\cite{hermans2017defense}, semi-hard negative pair mining~\cite{schroff2015facenet}, and soft-hard mining~\cite{yu2018hard}.
However, mining strategies could lead to a biased model because they usually account for a small selected minority and a large non-selected majority~\cite{wu2017sampling,schroff2015facenet,zheng2019hardness}.

To address this problem, recent works~\cite{duan2018deep,zhao2018adversarial,zheng2019hardness} have proposed using generative adversarial networks and autoencoders to generate synthetic hard samples.
These methods enable non-selected majorities to be exploited by synthesizing them into hard samples and training a model with augmented information.
Despite a performance boost of deep metric learning, it also suffers from several limitations.
First, along with a model for metric learning, an additional sub-network is required to generate synthetic hard samples, which leads to an increase in model size, hyper-parameters, and training time.
Moreover, deploying a generative model, such as a generative adversarial network, can result in optimization difficulty~\cite{arjovsky2017wasserstein}.

In this paper, we propose a simple yet powerful method for synthetic hard sample generation called symmetrical synthesis to address aforementioned limitations.
As illustrated in Figure~\ref{fig:teaser}, given two feature points within the same class, our proposed method generates symmetrical synthetic points with each other as an axis of symmetry.
% Given two feature points within the same class, our proposed method generates a synthetic point where an feature point is a symmetric with respect to a vector of the other feature point.
Then, it selects the hardest negative pair within the original and synthetic points.
This allows to train a model to pushes away samples of different classes with a stronger power.
In contrast to previous methods, as our method only requires simple algebraic computation to generate synthetic points, it is hyper-parameter free and can be applied to existing metric learning losses in a plug-and-play manner, without any modification of network architecture.
Furthermore, deploying our proposed method does not influence the training speed and optimization difficulty.
We demonstrate that deploying our proposed method gives a significant improvement in image clustering and retrieval tasks on CUB-200-2011, CARS196, and Standford Online Products.
The proposed method outperforms previous methods by wide margins.

\section{Related Work}

Our work is related to three lines of active research: (1) metric learning, (2) hard negative pair mining, and (3) hard sample generation.

\subsubsection{Metric Learning}
% History
% Classical metric learning methods find a better Mahalanobis distance~\cite{globerson2006metric,davis2007information} in linear space, which has limited capacity for modeling high-order correlations.
% Because deep learning allows a non-linear feature representation to be learned directly, metric learning losses have been proposed based on similarity and distance using the feature representation.
Metric learning losses have been proposed based on similarity and distance using the feature representation.
% Contrastive loss
% One of the simplest losses is the contrastive loss~\cite{chopra2005learning}, which takes pairs of samples and separates samples from different classes with a fixed margin, and pulls them closer otherwise.
% Triplet loss
One of the simplest losses is the triplet loss~\cite{weinberger2009distance}, which takes triplets of samples to separate the negative pair more than the positive pair with a fixed relative margin.
Despite its success, it has been reported to require expensive sampling methods to provide non-trivial samples for an efficient training~\cite{chechik2010large,cui2016fine}.
% N-pair loss
To address this problem, N-pair loss~\cite{sohn2016improved} is proposed to expand the idea of triplet loss by considering $N-1$ negative samples of different classes.
% Lifted-Struct loss
Similar to N-pair loss, lifted structure loss~\cite{oh2016deep} is proposed to train the embedding function by incorporating all negative samples within a batch.
% Angular loss
Angular loss~\cite{wang2017deep} considers that the distance metric is sensitive to scale and only considers second-order information between samples.
To circumvent these problems, angular loss constraints the angle at the negative point of triplet triangles.

\subsubsection{Hard Negative Pair Mining}
% History
% Hard negative pair mining has played an essential role in the performance of many machine learning tasks, such as deep metric learning, semi-supervised learning, and unsupervised learning.
Hard negative pair mining has played an essential role in the performance of deep metric learning.
The purpose of this strategy is to progressively select false positive samples, which can give more information in the training process.
% Offline hard negative pair mining
For example, offline hard negative pair mining~\cite{ahmed2015improved} is proposed to iteratively fine-tune a model with hard negative samples selected by a previously trained model.
% Hardest positive and negative triplet
Online hard negative pair mining~\cite{hermans2017defense} proposes the selection of the hardest positive and negative within a batch to compute the triplet loss.
% Semi-hard triplet
Semi-hard negative pair mining~\cite{schroff2015facenet} is proposed to avoid too confusing samples, such as the hardest positives and negatives, which may often be noise in data.
One of the limitations is that mining strategies usually focus on the selected minority and overlook the non-selected majority, which can lead to a biased model~\cite{wu2017sampling,schroff2015facenet,zheng2019hardness}.

\subsubsection{Hard Sample Generation}
Recently, there have been attempts to generate synthetic hard samples for exploiting a large number of easy negatives and training a model with extra semantic information.
For example, the deep adversarial metric learning (DAML) framework~\cite{duan2018deep} is proposed to generate synthetic hard samples from the easy negative samples in an adversarial manner.
Similarly, an adversarial network for hard triplet generation~\cite{zhao2018adversarial} is proposed to train a model with synthetic hard samples.
The hardness-aware deep metric learning (HDML) framework~\cite{zheng2019hardness} exploits an autoencoder architecture to generate label-preserving synthetics in the embedding space and manipulate their hard levels.
Nevertheless, all above-mentioned methods require additional generative networks, which result in a bigger model, slower training speed, and more hyper-parameters.
Our work re-defines the core component of generation by geometrical approach with simple algebraic computation in the embedding space instead of using the generative networks.
We show it can be easily used to existing metric learning losses without additional hyper-parameter, training speed decrease, and network modification.

\section{Proposed Method}

In this section, we present a novel method of synthetic hard sample generation called symmetrical synthesis (Symm).
% and its hard negative pair mining for metric learning losses.
% We first review deep metric learning and the standard similarity-based losses in a mathematical form.
As illustrated in Figure~\ref{fig:teaser}, the proposed method follows two steps: (1) symmetrical synthetic generation and, (2) hard negative pair mining.
% We first show how to generate symmetrical synthetics with an algebraic formulation.
% We then derive the combined loss between symmetrical synthetics and metric learning loss with hard negative pair mining.

\subsection{Symmetrical Synthesis}
\label{sec:symmetrical_synthesis}

\begin{figure}[t]
\centering
\includegraphics[width=0.6\columnwidth]{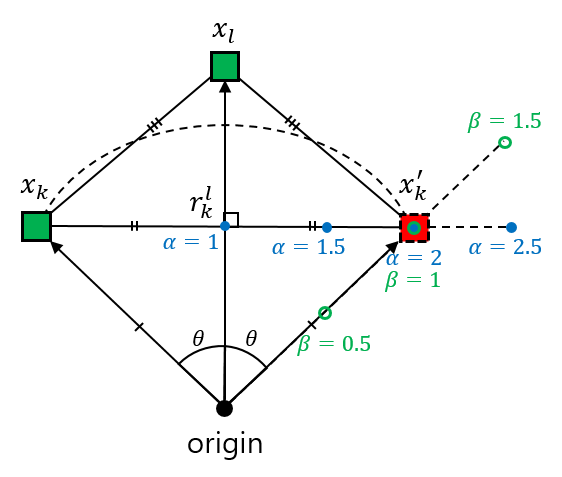} 
\caption{Illustration of generating symmetrical feature point. Green rectangles denote original feature points from the same class while a red with dotted boundary is synthetic feature points.}
\label{fig:symm_gen}
\end{figure}

% Symmetrical synthesizing formulation
The first step for the proposed method is to generate symmetrical synthetic points in the embedding space.
Let $\mathbfcal{I}$ be the data space and $\mathbfcal{X}$ be the $d$-dimensional embedding space.
We define $f:\mathbfcal{I} \xrightarrow{f} \mathbfcal{X}$ be the mapping from the data space to the embedding space parameterized by a deep neural network.
We sample a set of feature points $\mathbf{X} = [\mathbf{x}_1, \mathbf{x}_2, \dots, \mathbf{x}_N]$ where each point $\mathbf{x}_i$ has label $l_i \in \{1, \dots, C\}$.
%%%%
% 설명의 편의를 위해 우리는 mini-batch 안의 class i에 대한 두 개의 feature points를 ($\mathbf{x}_i^1$, $\mathbf{x}_i^2$)로 표현하였다.
%%%%
As illustrated in Figure~\ref{fig:symm_gen}, given two feature points ($\mathbf{x}_k$, $\mathbf{x}_l$) from the same class, synthetic points ($\mathbf{x}_k^{'}$, $\mathbf{x}_l^{'}$) can be generated with each other as an axis of symmetry.
In order to get $\mathbf{x}_k^{'}$, we define $\mathbf{r}_k^l$, which is a projection of $\mathbf{x}_k$ onto $\mathbf{x}_l$, i.e.,
\begin{align}
\mathbf{r}_k^l=\big(\mathbf{x}_k\cdot\mathbf{u}_{\mathbf{x}_l}\big)\mathbf{u}_{\mathbf{x}_l},
\end{align}
where $\mathbf{u}_{\mathbf{x}_l}$ is an unit vector of $\mathbf{x}_l$:
$\mathbf{u}_{\mathbf{x}_l} = \nicefrac{\mathbf{x}_l}{\|\mathbf{x}_l\|}$.
The synthetic point $\mathbf{x}_k^{'}$ is represented with a simple algebraic formulation as:
\begin{align}
\mathbf{x}_k^{'} & = \beta\big[\alpha\big(\mathbf{r}_k^l - \mathbf{x}_k\big) + \mathbf{x}_k\big],
\label{eq:synthetic}
\end{align}
where $\alpha$ is for how far the synthetic point is from the original point and $\beta$ is for how large the norm of synthetic point is.
The symmetrical synthetic point can be obtained when $\alpha =2.0$ and $\beta =1.0$.
Note that $\alpha$ and $\beta$ are only for explanation and an experiment, and they are not hyper-parameters.
The other symmetrical synthetic of $\mathbf{x}_l^{'}$ can be generated the same way.
Then we will obtain four feature points: two original and two synthetic.

There are two reasons why synthetic points should be generated with symmetric property.
The first is that symmetrical synthesis gives the same cosine similarity and Euclidean distance among pairs ($\mathbf{x}_k$ $\leftrightarrow$ $\mathbf{x}_l$ = $\mathbf{x}_k^{'}$ $\leftrightarrow$ $\mathbf{x}_l$ = $\mathbf{x}_k$ $\leftrightarrow$ $\mathbf{x}_l^{'}$).
This allows the generated points will not affect the positive side of the loss because any positive point included in a selected negative pair will have the same similarity and distance as described in Figure~\ref{fig:pos_syn}.
% This allows the loss to push negative pairs more effectively without affecting the positive side of the loss.
The second reason is that the generated synthetic point will always have the same norm as the original point.
Every metric learning loss can be influenced by the norm.
To control it, triplet loss conducts $l2$-normalization to project feature points onto hyper-sphere space~\cite{weinberger2009distance}, while N-pair and angular loss regularize the norm without $l2$-normalization in Euclidean space~\cite{sohn2016improved,wang2017deep}.
Thus, a synthetic point generated by an $l2$-normalized point would lie in hyper-sphere space, and a synthetic point generated by a non-$l2$-normalized point would have the same norm as the original point in Euclidean space.
This gives continuity of control over the norm during the training process and does not disturb optimization.

\subsection{Metric Learning with Symmetrical Synthesis}

\begin{figure}[t]
\centering
\includegraphics[width=0.6\columnwidth]{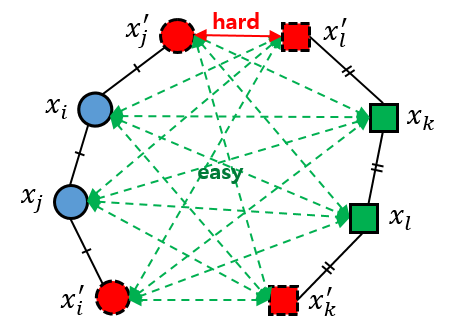} 
\caption{Possible negative pairs between two different classes including symmetrical synthetics. Rectangles and circles represent two different classes. Green and blue are original feature points while red with dotted boundary are synthetic feature points.}
\label{fig:pos_syn}
\end{figure}

% Symm + Triplet, Symm + N-pair loss ...
To exploit the generated symmetrical synthetics, we perform hard negative pair mining for each metric learning loss.
Rather than taking negative pairs based on an anchor, as in Figure~\ref{fig:teaser}, we further use all original and synthetic points from the positive class to enlarge the number of negative pairs, as in Figure~\ref{fig:pos_syn}.
Given four feature points ($\mathbf{x}_i$, $\mathbf{x}_j$, $\mathbf{x}_i^{'}$, $\mathbf{x}_j^{'}$) from a positive class and ($\mathbf{x}_k$, $\mathbf{x}_l$, $\mathbf{x}_k^{'}$, $\mathbf{x}_l^{'}$) from a negative class, we first compute the similarities of the 16 possible negative pairs between positive and negative points.
Then, we select the hardest negative pair for metric learning loss.
Because the cosine similarity and Euclidean distance of pairs are same by symmetric property ($\mathbf{x}_i$ $\leftrightarrow$ $\mathbf{x}_j$ = $\mathbf{x}_i^{'}$ $\leftrightarrow$ $\mathbf{x}_j$ = $\mathbf{x}_i$ $\leftrightarrow$ $\mathbf{x}_j^{'}$), we use the original positive points for positive pair (i.e., $\mathbf{x}_i$ $\leftrightarrow$ $\mathbf{x}_j$) for simplicity.
We formulate combinations of symmetrical synthesis with existing metric learning losses.

Let $\mathcal{P}$ be a set of positive pairs with original points and $\widehat{\mathcal{N}}_{l_i,l_k}$ be a set of negative pairs with a positive point from the class $l_i$ and a negative point from the class $l_k$ including symmetrical synthetics.
Triplet loss considers triplets of samples defined as:
%% Triplet
\begin{align}
%vanilla triplet
\mathcal{L}_{triplet} = \frac{1}{|\mathcal{P}|}\sum_{\substack{(i, j)\in{\mathcal{P}} \\ k:l_i \neq l_k}}\Big[ D_{i,j}^2 - D_{i,k}^2 + m \Big]_+,
\end{align}
where $m$ is a margin, $D_{i,j} = \| \mathbf{x}_i - \mathbf{x}_j \|_2$ is the Euclidean distance, and $[\cdot]_+$ denotes the hinge function~\cite{weinberger2009distance}.
For symmetrical synthesis, we combine hard negative pair mining with triplet loss by min-pooling among Euclidean distances of negative pairs in $\widehat{\mathcal{N}}_{l_i,l_k}$:
\begin{align}
\mathcal{L}_{triplet}^{Symm} = \frac{1}{|\mathcal{P}|}\hspace*{-0.1cm}\sum_{\substack{(i, j)\in\mathcal{P} \\ k:l_i \neq l_k}}\hspace*{-0.15cm}\Big[ D_{i,j}^2 - \hspace*{-0.2cm}\min_{\substack{(p, n)\in{\widehat{\mathcal{N}}_{l_i,l_k}}}} \hspace*{-0.2cm} D_{p,n}^2 + m \Big]_+,
\end{align}

Lifted structure loss compares the distances against all negative pairs for each positive pair and pushes all negative points farther than a margin.
More precisely, it minimizes
%%% lifted-structure
\begin{align}
\mathcal{L}_{lifted} & = \frac{1}{2|\mathcal{P}|} \sum_{(i,j) \in \mathcal{P}}\bigg[\log\bigg\{ \sum_{k:l_i \neq l_k}\exp\big(m-D_{i,k}\big) \nonumber \\
& + \sum_{k:l_j \neq l_k}\exp\big(m - D_{j,k} \big)\bigg\} + D_{i,j} \bigg]^2_+,
\end{align}
Similarly to the triplet loss, the combination of symmetrical synthesis and lifted structure loss can be formulated by using min-pooling as follows:
\begin{align}
\mathcal{L}_{lifted}^{Symm} & = \frac{1}{|\mathcal{P}|} \sum_{(i,j) \in \mathcal{P}}\bigg[\log\bigg\{ \nonumber \\
& \hspace*{-0.5cm}\sum_{k:l_i \neq l_k}\exp\big(m-\min_{(p,n)\in\widehat{\mathcal{N}}_{l_i,l_k}}{D_{p,n}}\big)\bigg\} + D_{i,j} \bigg]^2_+,
\end{align}

For N-pair loss, additional negative samples are considered into triplets, and the triplet is turned into an N-tuplet.
% For N-pair loss, it is trained with additional negative samples into triplets and turns the triplet into an N-tuplet.
The loss is defined as:
%%%%%%%%%%%%%%%%%%% npair
%% vanilla npair
\begin{align}
\mathcal{L}_{npair} = \frac{1}{|\mathcal{P}|} \sum_{(i,j) \in \mathcal{P}}\bigg\{ \log\bigg[ 1 + \sum_{k:l_i \neq l_k}\exp\big(S_{i,k} - S_{i,j} \big)\bigg] \bigg\},
\end{align}
where $S_{i,j}={\mathbf{x}_i}^T \mathbf{x}_j$ is the similarity between embedding $\mathbf{x}_i$ and $\mathbf{x}_j$.
We formulate N-pair loss with symmetrical synthesis by adding max-pooling because of cosine similarity, and perform hard negative pair mining on every negative class in a mini-batch:
%% symm-npair
\begin{align}
\mathcal{L}_{npair}^{Symm} = &\frac{1}{|\mathcal{P}|} \sum_{(i,j) \in \mathcal{P}}\bigg\{ \log\bigg[ 1 \nonumber \\
& + \sum_{k:l_i \neq l_k}\exp\big(\max_{(p,n)\in\widehat{\mathcal{N}}_{l_i,l_k}}S_{p,n} - S_{i,j} \big)\bigg] \bigg\},
\end{align}

%%% angular
Angular loss is proposed to encode the third-order relation to triplet in terms of the angle at the negative point:
\begin{align}
&\mathcal{L}_{ang} = \frac{1}{|\mathcal{P}|} \sum_{(i,j) \in \mathcal{P}}\bigg\{ \log\bigg[ 1 + \sum_{k:l_i \neq l_k}\exp\big(f_{i,j,k}^n - f_{i,j}^p \big)\bigg] \bigg\},
\end{align}
where $f_{i,j}^p = 2(1+\tan^2\alpha){\mathbf{x}_i}^T\mathbf{x}_j$ and $f_{i,j,k}^n = 4\tan^2\alpha(\mathbf{x}_i + \mathbf{x}_j)^T\mathbf{x}_k$.
Similarly to the N-pair loss, we can combine the symmetrical synthesis with angular loss by adding max-pooling for hard negative pair mining on every negative class as follows:
\begin{align}
\mathcal{L}_{ang}^{Symm} = & \frac{1}{|\mathcal{P}|} \sum_{(i,j) \in \mathcal{P}}\bigg\{ \log\bigg[ 1 \nonumber \\
& \hspace*{-0.5cm} + \sum_{k:l_i \neq l_k}\exp\big(\max_{(p,q,r)\in \widetilde{\mathcal{N}}_{l_i,l_k}} f_{p,q,r}^n - f_{i,j}^p \big)\bigg] \bigg\},
\end{align}
where $\widetilde{\mathcal{N}}_{l_i,l_k}$ is the set of triplets with two positive points from the class $l_i$ and one negative point from the class $l_k$ which is utilized in $f_{i,j,k}^n$.

% Effect of symmetrical synthesis
Metric learning with the proposed symmetrical synthesis has two effects.
First, using synthetic feature points leads to a more generalized model, because trivial samples, which could have been ignored by mining strategies, can be exploited by generating synthetic points and training the model with augmented information.
Secondly, hard negative pair mining within the original and synthetic points allows metric learning losses to push away between different classes with greater force.
This leads to higher inter-class variation with better clustering in the embedding space.

\section{Experiments}

In this section, we report experimental results from the proposed symmetrical synthesis on both image clustering and retrieval tasks.
% We conduct experiments to analyze the effect of the proposed method and evaluate its quantitative and qualitative performance.
To evaluate quantitative performance, we use the standard $F_1$ and NMI metrics~\cite{manning2010introduction} for the image clustering task, and Recall@K score for the image retrieval task.

\subsection{Datasets}

We evaluate our proposed method on the widely used three benchmarks by following the conventional protocol of train and test splits used by~\cite{zheng2019hardness,oh2016deep}.
(1) CUB-200-2011 (CUB200)~\cite{wah2011caltech} has 11,788 images of 200 bird species, where the first 5,864 images of 100 species are used for training and the remaining 5,924 images of 100 species are used for testing.
(2) CARS196~\cite{krause20133d} has 16,185 car images of 196 classes. We use the first 8,054 images of 98 classes for training and the remaining 8,131 images of 98 classes for testing.
(3) Stanford Online Products (SOP)~\cite{oh2016deep} datasets contains 120,053 product images of 22,634 classes, where the first 59,551 images of 11,318 classes are used for training and the remaining 60,502 images of 11,316 classes are used for testing.
For CUB200 and CARS196, our method is evaluated without the bounding box information.

\subsection{Experimental Setting}
Throughout the experiments, TensorFlow~\cite{abadi2016tensorflow} framework is used on a Tesla P40 GPU with 24GB memory.
All images are normalized to 256 $\times$ 256, horizontal flipped and randomly cropped to 227 $\times$ 227.
The embedding size is set to 512-dimensional for all feature vectors.
Triplet and lifted structure loss use $l2$-normalized features with Euclidean distance, and N-pair and angular loss use non-$l2$-normalized features with cosine similarity.
We use ImageNet~\cite{deng2009imagenet} pre-trained GoogLeNet~\cite{szegedy2015going} and the Xavier method~\cite{glorot2010understanding} to random initialize a fully connected layer.
We set the learning rate to $10^{-4}$ with the Adam optimizer~\cite{kingma2014adam}.
The batch size of 128 is used for every dataset.

\subsection{Experimental Results}
We perform experiments to analyze the effect of our proposed method.
The following experiments are conducted on the CARS196 dataset with N-pair loss in the image clustering and retrieval task.
% We used the proposed symmetrical synthesis considering positive synthesis unless otherwise noted in the experiment.

\begin{figure}[t!h!]
\centering
\includegraphics[width=0.9\columnwidth]{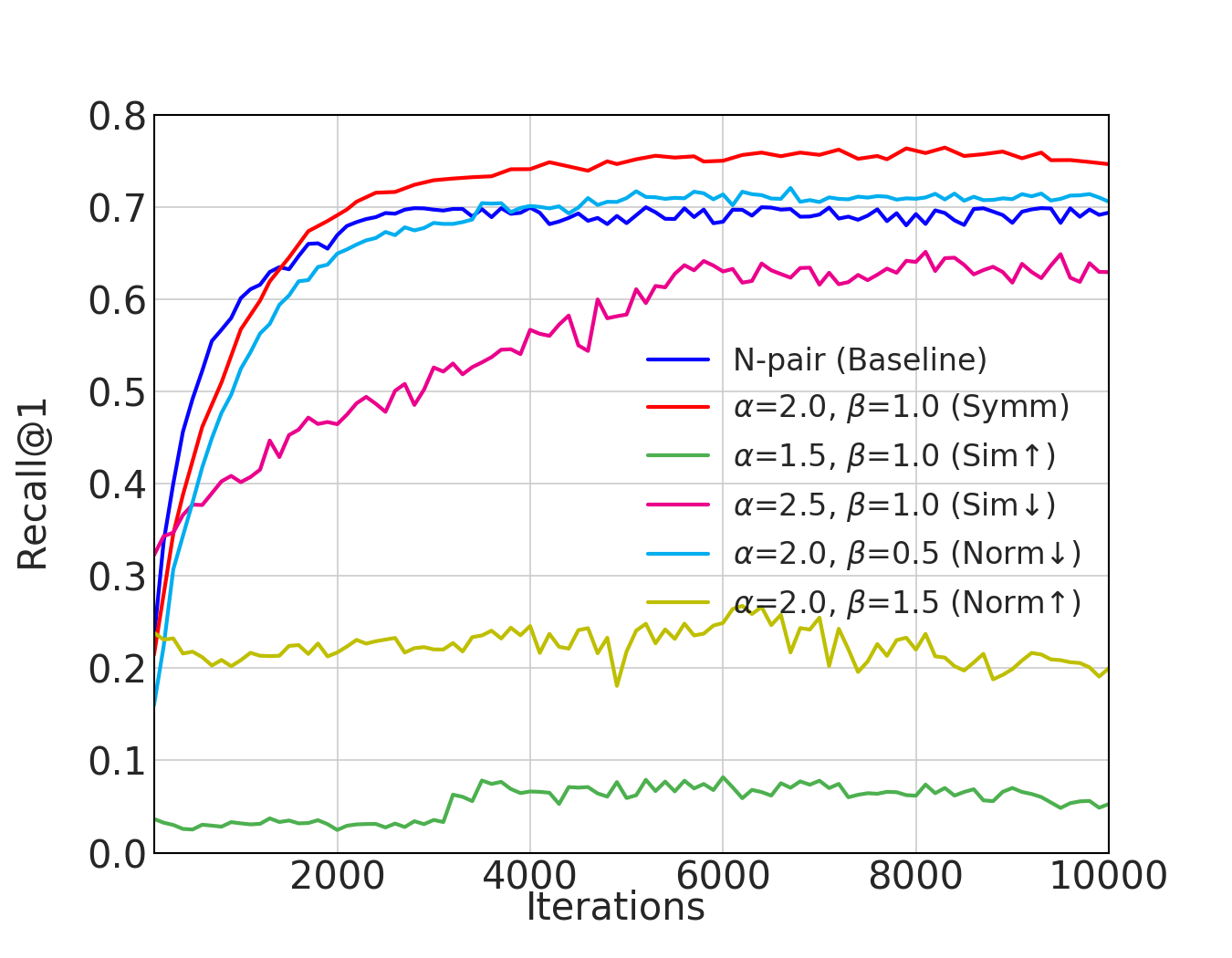} 
\caption{Recall@1 curve for comparison of different similarity and norm.
The baseline is N-pair loss, while the rest is Symm + N-pair loss with different $\alpha$ and $\beta$ on CARS196 dataset.}
\label{fig:alpha_beta}
\end{figure}

\begin{figure}[t!h!]
\centering
\includegraphics[width=0.9\columnwidth]{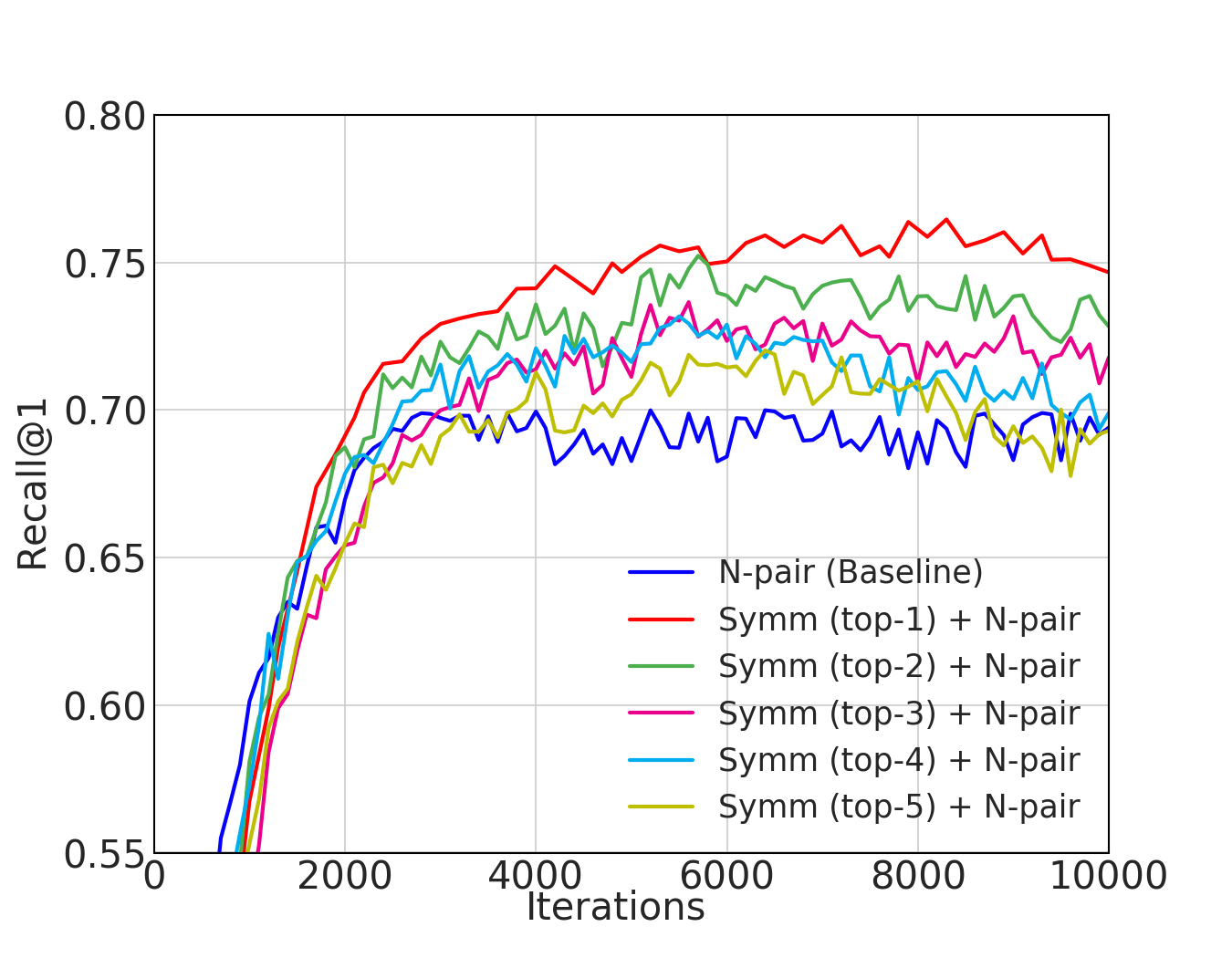} 
\caption{Recall@1 curve for comparison of top-$k$ hard negative pair mining.
It is trained and evaluated with N-pair loss for baseline and Symm + N-pair loss on CARS196 dataset.}
\label{fig:top_recall}
\end{figure}

% \begin{figure}[t!h!]
% \centering
% \includegraphics[width=0.9\columnwidth]{top_nmi.png} 
% \caption{NMI curve for comparison of top-$k$ hard negative pair mining in clustering task.
% It is trained and evaluated with N-pair loss for baseline and Symm + N-pair loss on CARS196 dataset.}
% \label{fig:top_nmi}
% \end{figure}

\begin{figure}[t!h!]
\centering
\includegraphics[width=0.9\columnwidth]{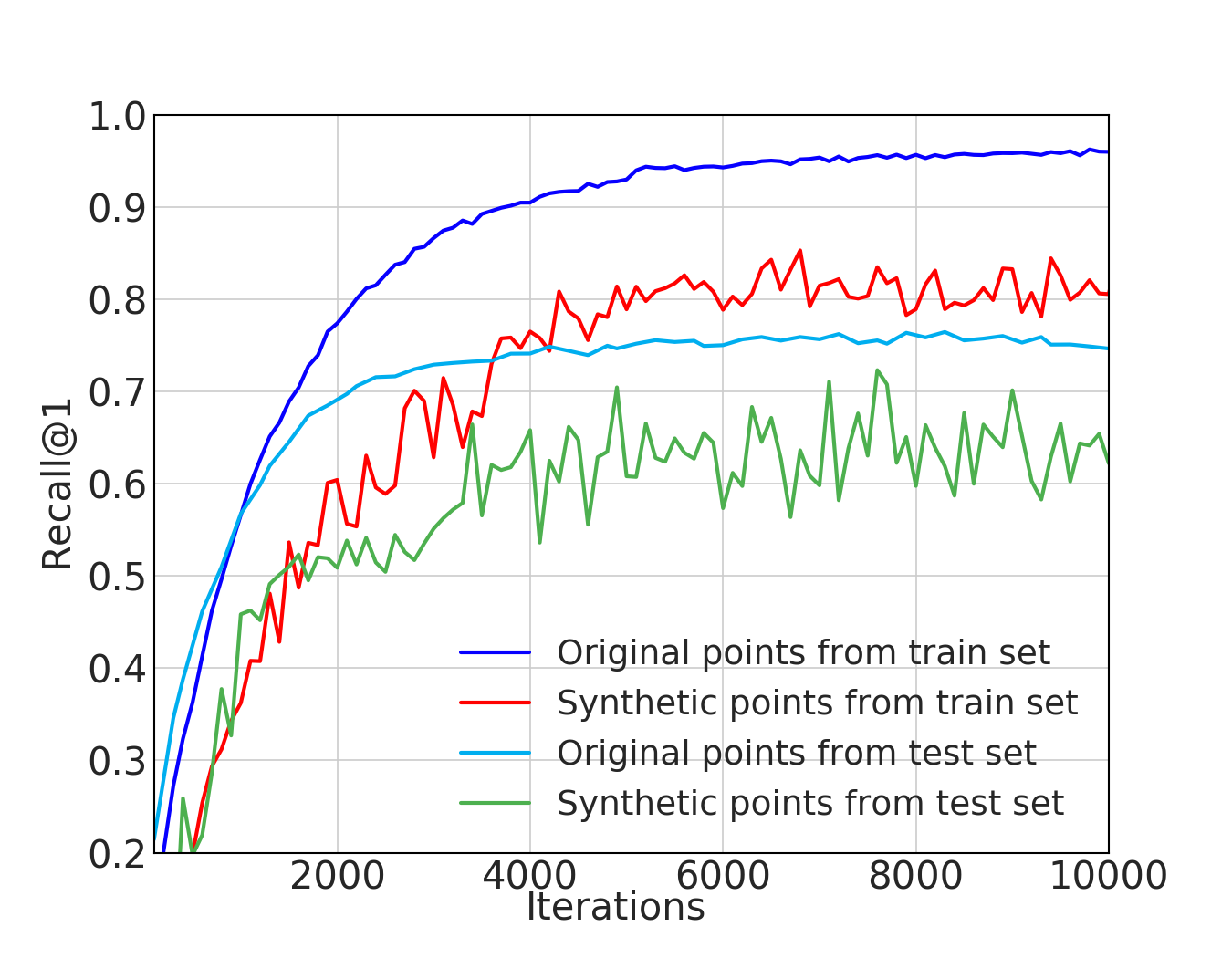} 
\caption{Recall@1 curve for comparison of original points and synthetic points from train and test set.
It is trained and evaluated with Symm + N-pair loss on CARS196 dataset.}
\label{fig:symm_recall}
\end{figure}

\begin{figure}[t!h!]
\centering
\includegraphics[width=0.9\columnwidth]{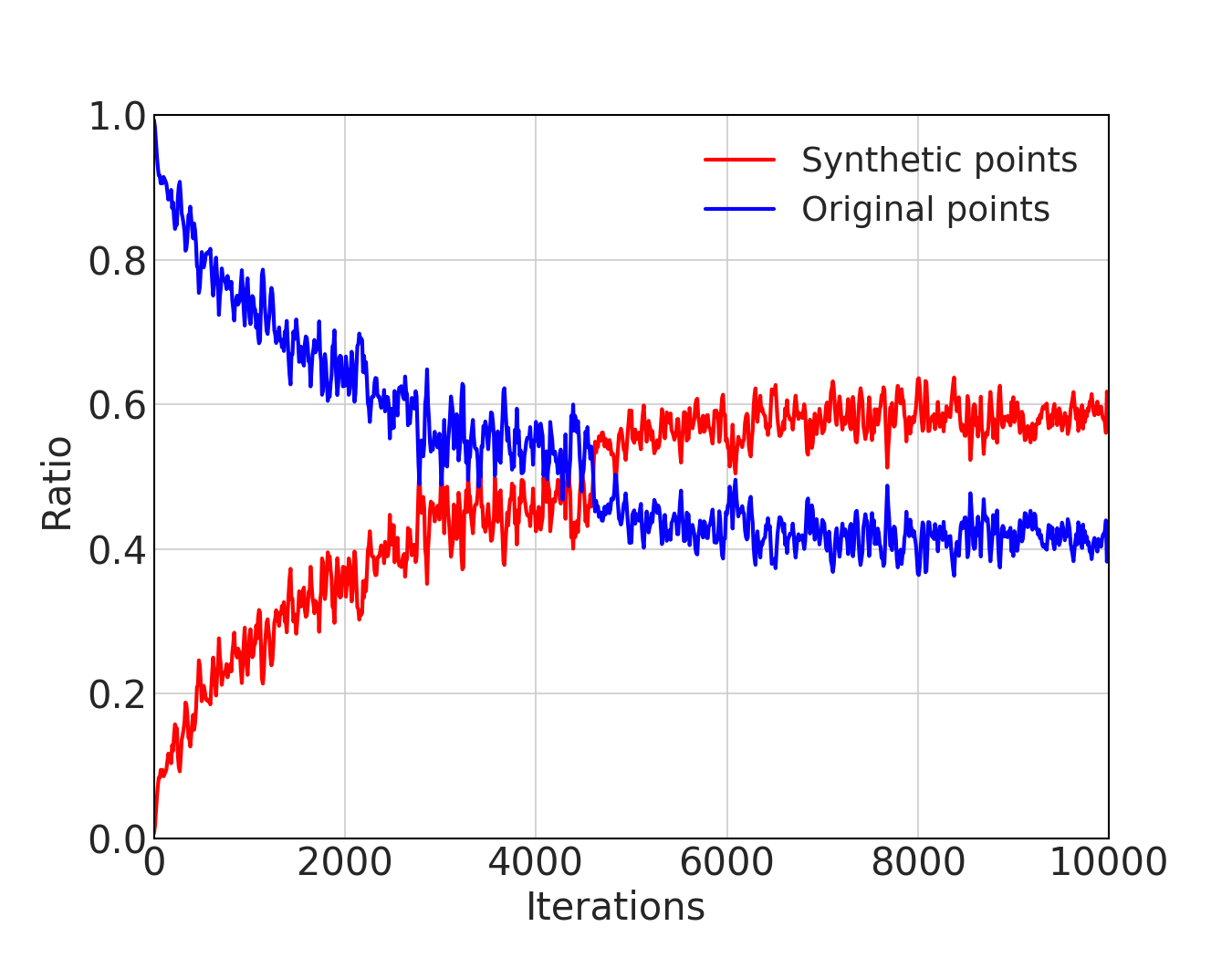} 
\caption{Ratio of selected feature points during hard negative pair mining between original and synthetic points.
It is trained with Symm + N-pair loss on CARS196 dataset.}
\label{fig:symm_ratio}
\end{figure}

\subsubsection{Impact of Similarity and Norm}
As mentioned in Section~\ref{sec:symmetrical_synthesis}, we generate synthetics with symmetric property so that the similarity and norm can be maintained.
To see the impact of similarity and norm, we conduct experiments by differentiating $\alpha$ and $\beta$ in Eq.~\ref{eq:synthetic}.
As illustrated in Figure~\ref{fig:symm_gen}, differentiating $\alpha$ gives a point with different cosine similarity and norm, but we force the norm to be same with the original point by multiplying $\nicefrac{\|x_k\|}{\|x_k^{'}\|}$ for the experiments.
With the same norm, larger cosine similarity ($\alpha =1.5$, $\beta =1.0$) is not trainable and smaller cosine similarity ($\alpha =2.5$, $\beta =1.0$) results in dramatic performance reduction.
Differentiating $\beta$ gives a point with the same cosine similarity and different norm.
With the same cosine similarity, larger norm ($\alpha =2.0$, $\beta =1.5$) is not trainable and smaller norm ($\alpha =2.0$, $\beta =0.5$) shows similar performance with N-pair, but lower than the proposed symmetrical synthetics ($\alpha =2.0$, $\beta =1.0$).
This demonstrates that maintaining similarity and norm by generating symmetrical synthetics is essential for network optimization and converged performance.

\begin{figure*}[t!h!]
     \centering
     \begin{subfigure}[b]{0.33\linewidth}
         \centering
         \includegraphics[width=1.0\columnwidth]{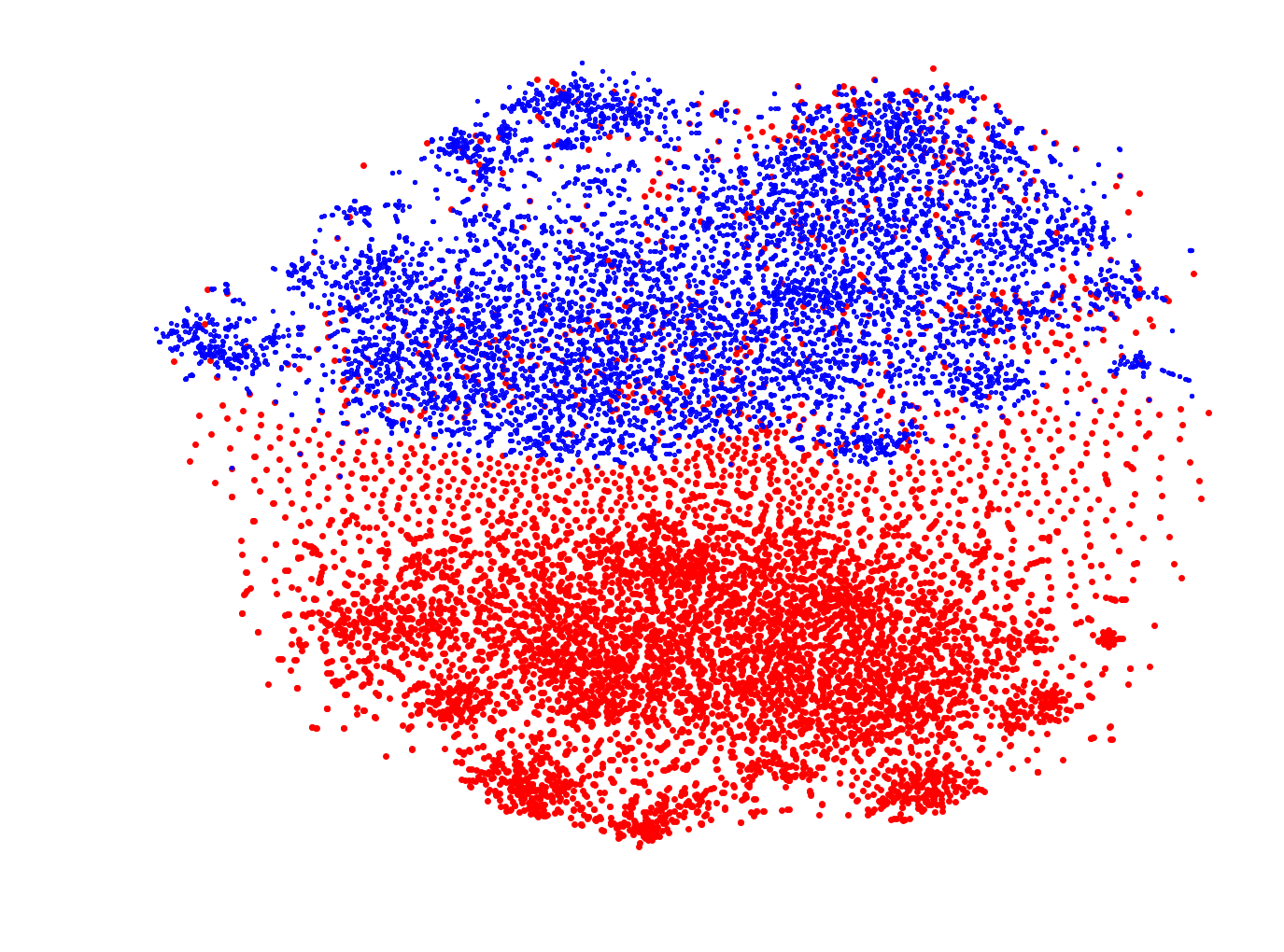}
         \caption{100 iterations}
         \label{fig:syn_100}
     \end{subfigure}
     \begin{subfigure}[b]{0.33\linewidth}
         \centering
         \includegraphics[width=1.0\columnwidth]{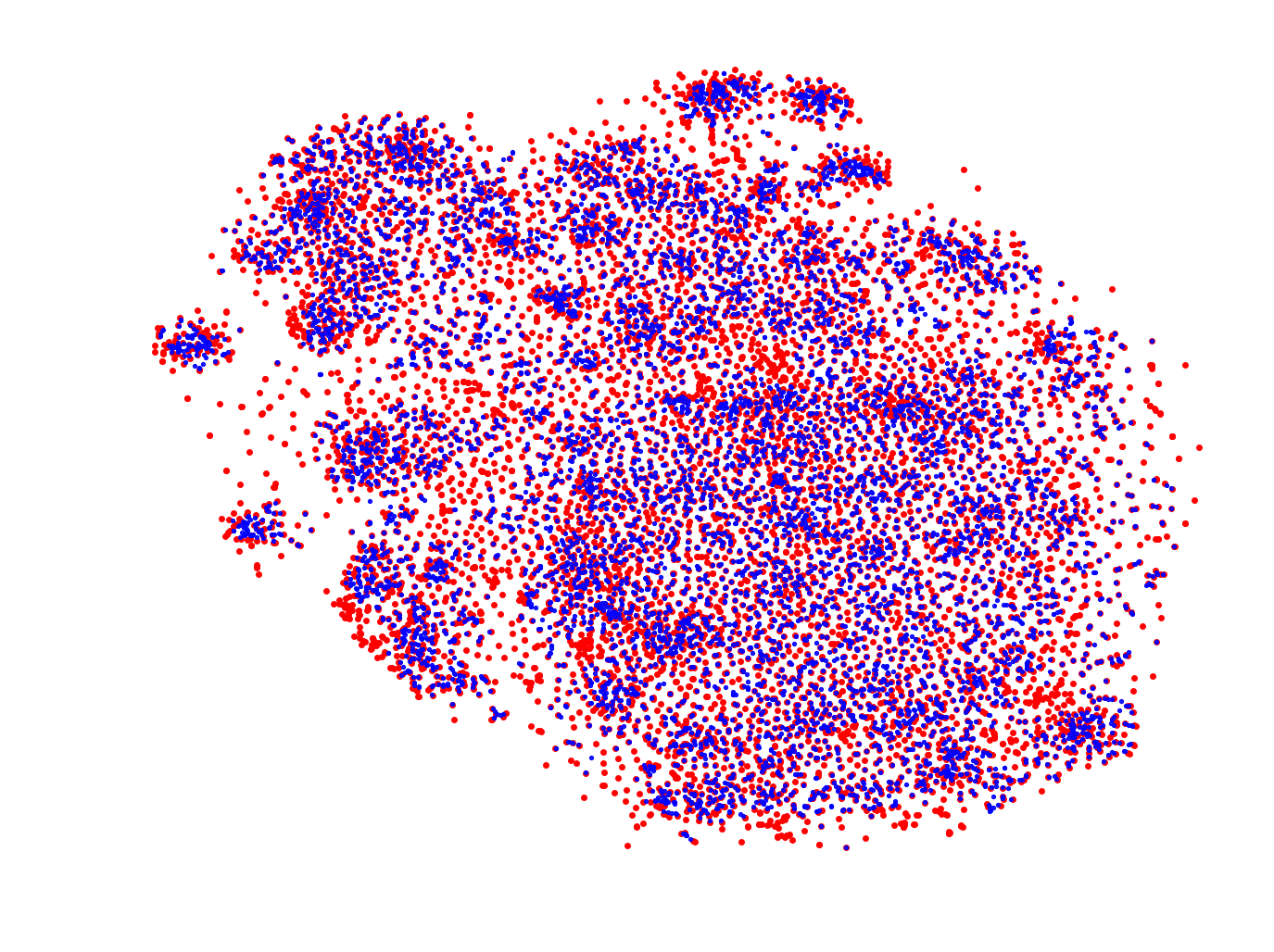}
         \caption{1000 iterations}
         \label{fig:syn_1000}
     \end{subfigure}
     \begin{subfigure}[b]{0.33\linewidth}
         \centering
         \includegraphics[width=1.0\columnwidth]{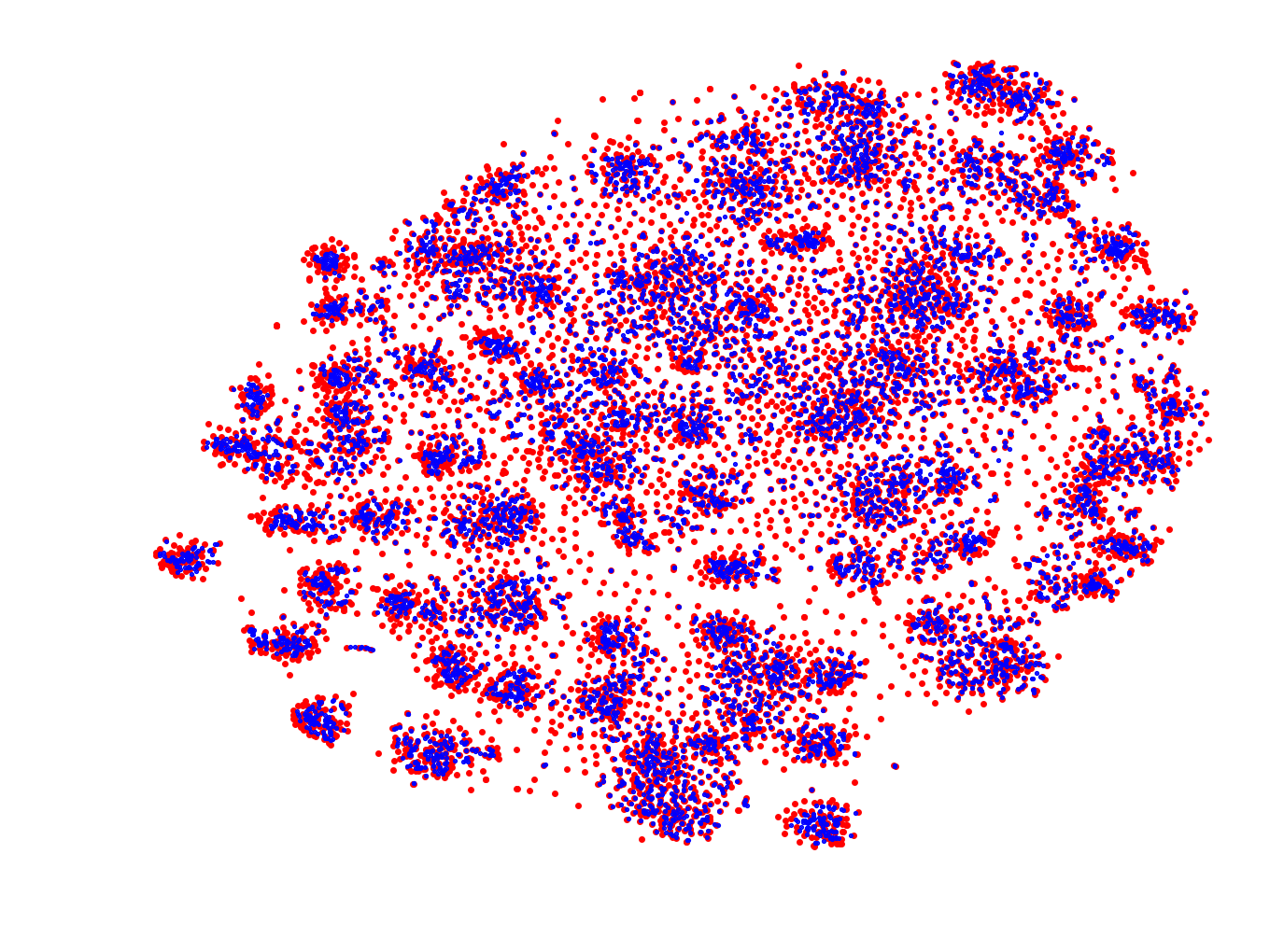}
         \caption{2000 iterations}
         \label{fig:syn_2000}
     \end{subfigure}
     \begin{subfigure}[b]{0.33\linewidth}
         \centering
         \includegraphics[width=1.0\columnwidth]{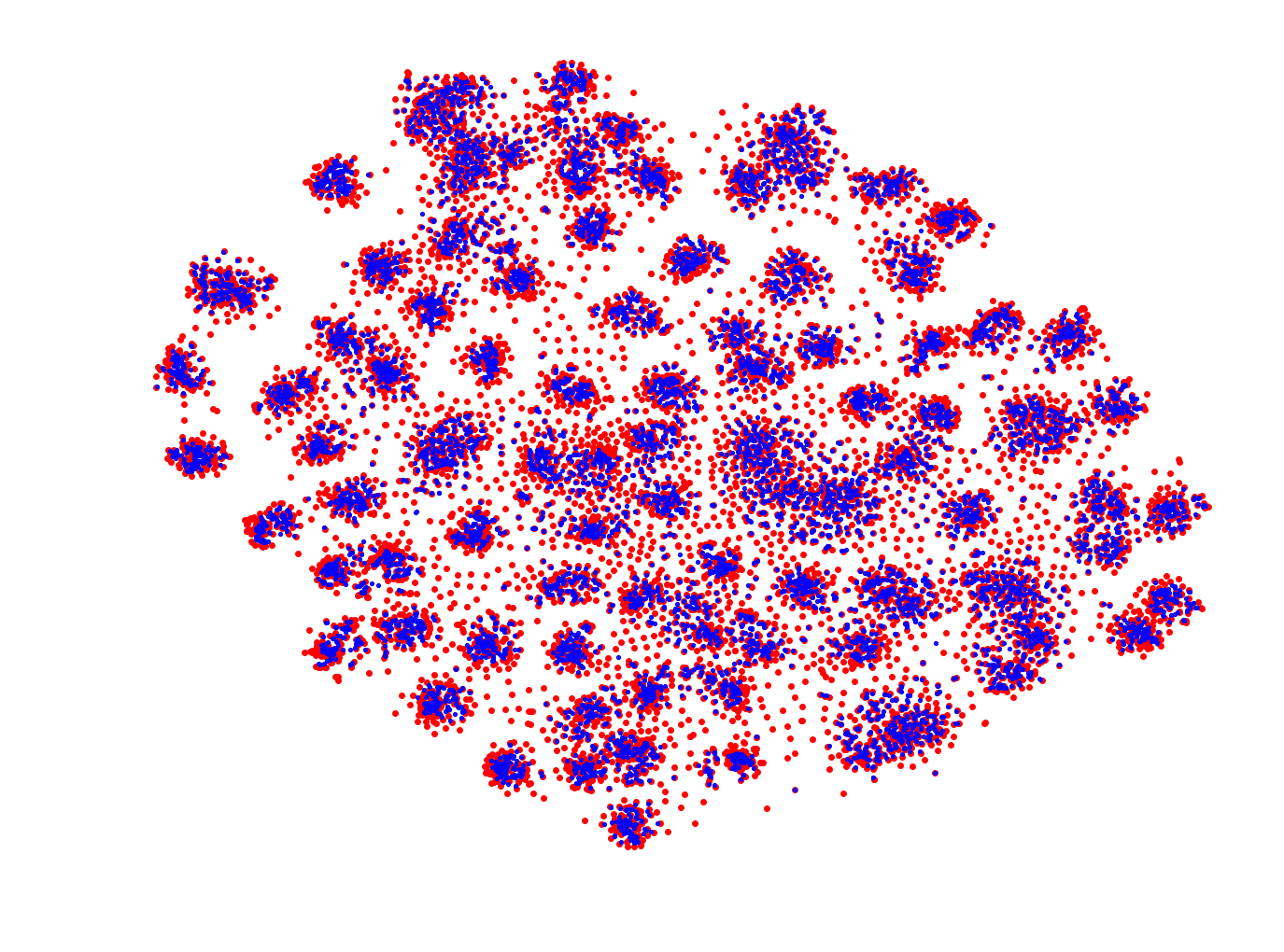}
         \caption{3000 iterations}
         \label{fig:syn_3000}
     \end{subfigure}
     \begin{subfigure}[b]{0.33\linewidth}
         \centering
         \includegraphics[width=1.0\columnwidth]{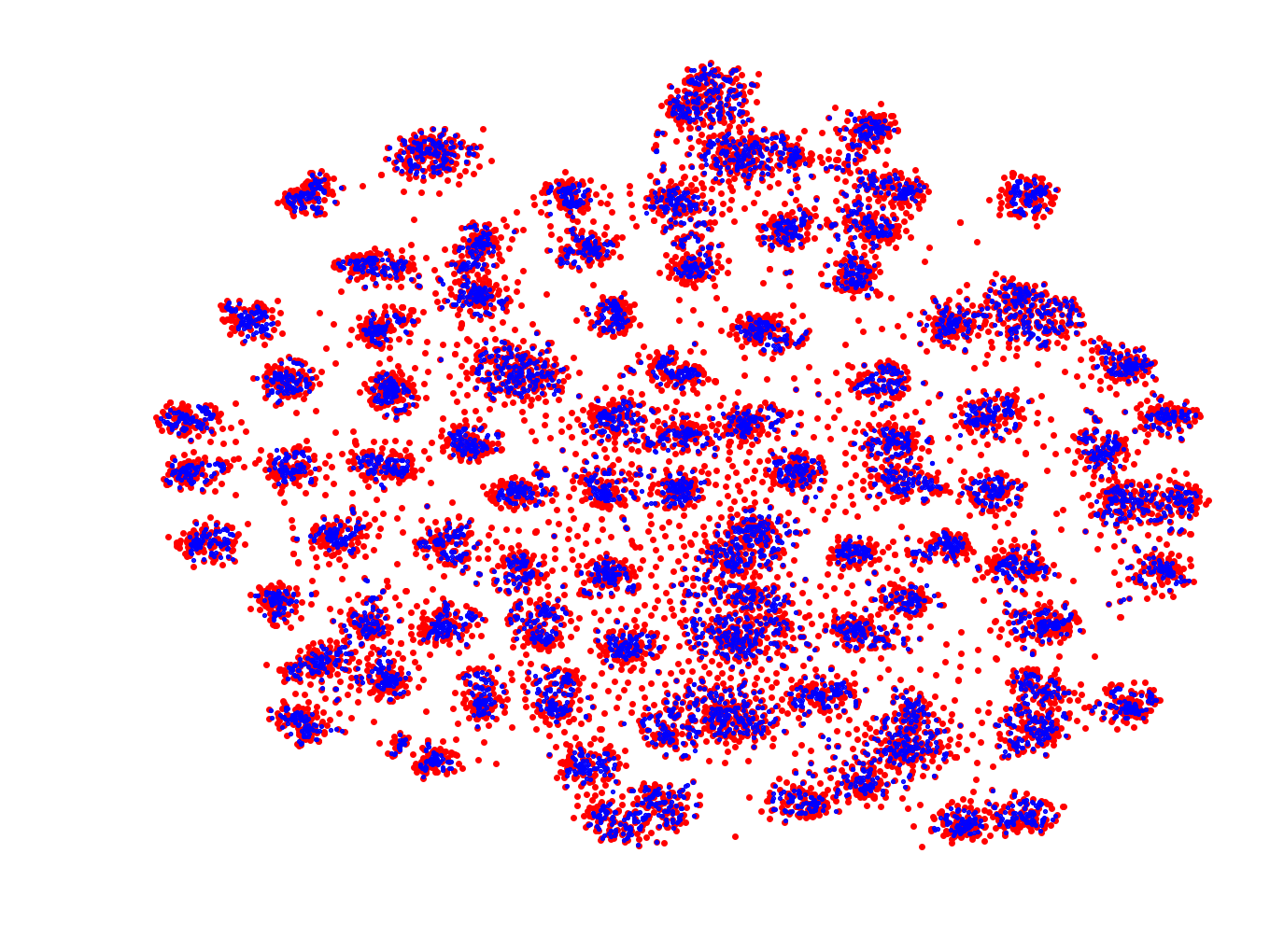}
         \caption{4000 iterations}
         \label{fig:syn_4000}
     \end{subfigure}
     \begin{subfigure}[b]{0.33\linewidth}
         \centering
         \includegraphics[width=1.0\columnwidth]{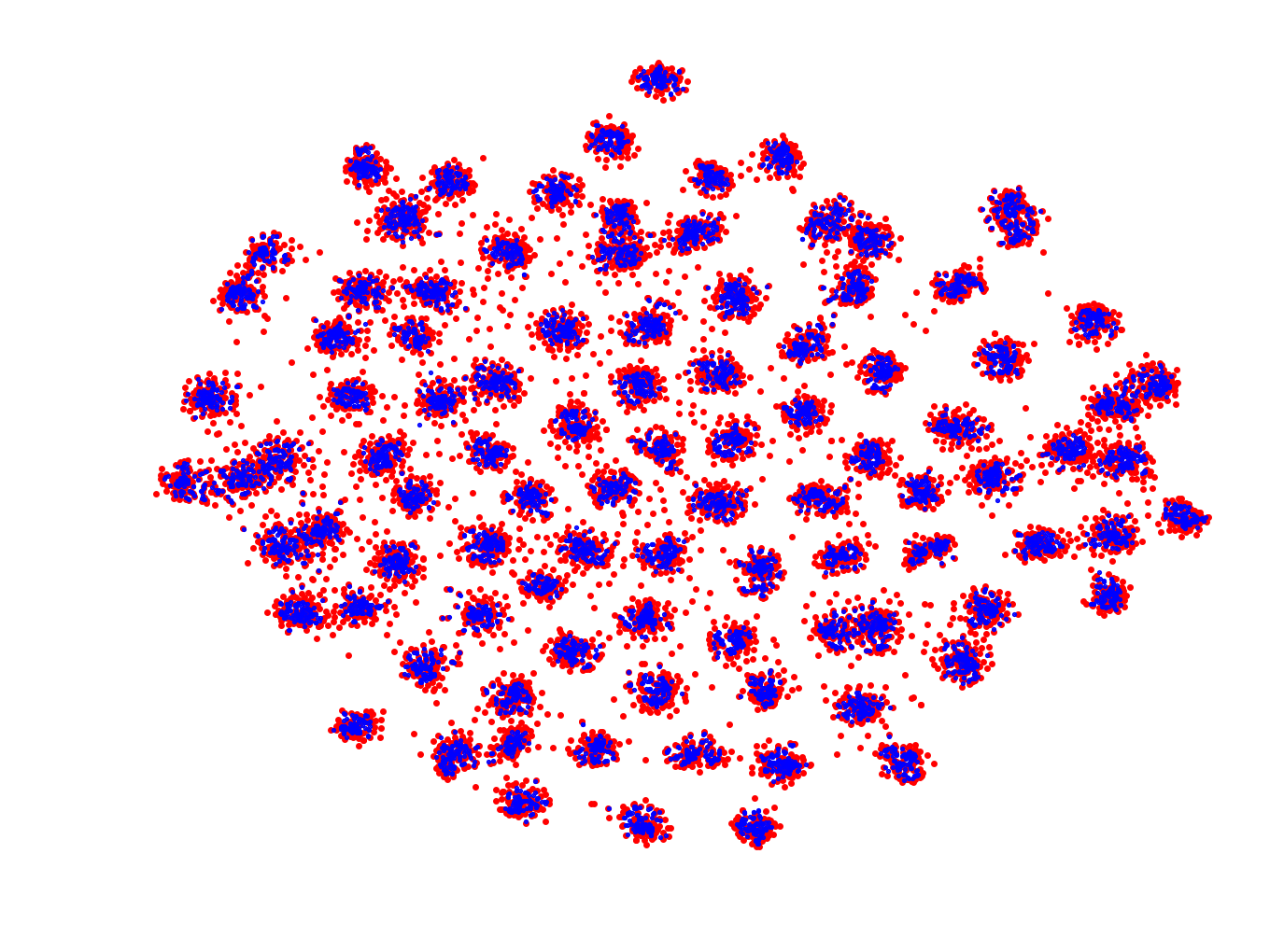}
         \caption{8000 iterations}
         \label{fig:syn_8000}
     \end{subfigure}
\caption{A t-SNE visualization of Symm + N-pair loss with the original (blue) and symmetrical synthetic (red) feature points from the training set of CARS196 dataset.}
\label{fig:tsne}
\end{figure*}

\subsubsection{Level of Hardness}
% Top-1,2,3,4,5,n-pair 성능 비교 (R@1, NMI)

In the proposed symmetrical synthesis, the similarity of every possible negative pair is computed and the hardest negative pair is selected.
This hard negative pair mining strategy is designed to use the most informative pairs for training.
To analyze the effect of the hard negative pair mining, we conduct experiments to compare among the proposed methods with different top-$k$ hardest negative pair mining with N-pair loss and the baseline model with only N-pair loss.
Figure~\ref{fig:top_recall} shows the learning curves of each model setting in the retrieval task.
We observe that the harder the pair selected, the higher the performance, where every model with the proposed method outperforms the baseline model in both tasks.
This is because harder pairs are more informative; thus, using the symmetrical synthesis with harder pairs pushes different classes away from each other with stronger power.

\subsubsection{Label of Synthetics}
% Synthetic point가 어느 부근에 생성되는지와 어떤 성능을 보이는지

As shown in Figure~\ref{fig:symm_recall}, we conduct experiments to estimate where the synthetic points are generated.
We evaluate the recall performance of the original points and synthetic points from the training and test sets.
Synthetic points are generated from the original points with randomly selected points in the same class as the axis of symmetry.
We speculate that synthetic points do not have to be inside the same cluster because they will only be used as hard samples to push the other class with stronger power.
However, we expect the synthetic points would lie around the boundary of the same cluster to work as hard samples.
We observe that the curve of synthetic points has more fluctuation than that of the original points in both the training and test sets.
Besides, the performance of the original points is always higher than that of the synthetic points.
We believe this is because a high portion of synthetic points is generated around the boundary of the cluster working as hard samples.
Also, the performance increase of synthetic points indicates that synthetic points lie in meaningful spots due to the increased clustering ability of the model.

\subsubsection{Visualization and Ratio of Feature Points}

The ideal place of generated symmetrical synthetic points is around the boundary of the class cluster so that the synthetic points can work as hard feature points during hard negative pair mining.
To see the geographical location of the symmetrical synthetic points compared to the original points, we visualize the embedding space of each training step with the Barnes-Hut t-SNE~\cite{van2014accelerating}, as shown in Figure~\ref{fig:tsne}.
Moreover, we conduct an experiment to see how many original and synthetic feature points are selected to be the hardest negative pair in each training step, as illustrated in Figure~\ref{fig:symm_ratio}.
The ratio of synthetic points is calculated as $ratio^{(syn)} = \frac{\#\,of\,synthetic}{\#\,of\,original\,+\,\#\,of\,synthetic}$, while the ratio of original points is calculated as $ratio^{(ori)} = 1 - ratio^{(syn)}$.

At the beginning of the training, original feature points are scattered without forming clusters, and the similarities of positive pairs will be relatively small, as shown in Figure~\ref{fig:syn_100}.
This causes the symmetrical synthetic points are generated on the meaningless place far from the positive pairs, that will be hardly selected during hard negative pair mining.
Hence, the original points are selected mostly over the synthetic points at first, as illustrated in Figure~\ref{fig:symm_ratio}.
The better clustering ability the model has, the higher the chance that the synthetic points will be generated around the boundary of the cluster to become the hard feature points.
Generated synthetic points start lying around the boundary of the class cluster from the 3000 steps, as shown in Figure~\ref{fig:syn_3000}.
After the 4000 steps, more than half of synthetic points are selected over the original points during hard negative pair mining, as illustrated in Figure~\ref{fig:symm_ratio}.
These selected synthetic points will work as hard negatives to train the model with richer information.
Finally, we obtain clean and well-clustered embeddings of the original feature points, as shown in Figure~\ref{fig:syn_8000}.
More details of visualization and ratio of feature points are given in the supplementary video.\footnote{\url{https://youtu.be/X9mJJKDokEU}}

\begin{table}[t!h!]
\centering
\resizebox{1.0\columnwidth}{!}{%
\begin{tabular}{c|cc|cccc}
\hline
\multirow{2}{*}{Method} & \multicolumn{2}{c|}{Clustering} & \multicolumn{4}{c}{Retrieval} \\ \cline{2-7} 
                        & NMI            & F1             & R@1   & R@2   & R@4   & R@8   \\ \hline\hline
% Contrastive             & 47.2           & 12.5           & 27.2  & 36.3  & 49.8  & 62.1  \\
% DDML                    & 47.3           & 13.1           & 31.2  & 41.6  & 54.7  & 67.1  \\ \hline
Triplet                 & 49.8           & 15.0           & 35.9  & 47.7  & 59.1  & 70.0  \\
Triplet$^\dagger$            & 53.4           & 17.9           & 40.6  & 52.3  & 64.2  & 75.0  \\
DAML (Triplet)          & 51.3           & 17.6           & 37.6  & 49.3  & 61.3  & 74.4  \\
HDML (Triplet)          & 55.1           & 21.9           & 43.6  & 55.8  & 67.7  & 78.3  \\
Symm+Triplet            & 59.6           & 26.2           & 51.4  & 63.0  & 74.4  & 84.1  \\
Symm+Triplet$^\dagger$  & \textbf{63.3} & \textbf{32.1} & \textbf{55.0} & \textbf{67.3} & \textbf{77.5} & \textbf{86.0}  \\ \hline
N-pair                  & 60.2           & 28.2           & 51.9  & 64.3  & 74.9  & 83.2  \\
DAML (N-pair)           & 61.3           & 29.5           & 52.7  & 65.4  & 75.5  & 84.3  \\
HDML (N-pair)           & 62.6           & 31.6  & 53.7  & 65.7  & 76.7  & 85.7  \\
Symm+N-pair             & \color{red}\textbf{63.6}  & \color{red}\textbf{32.5} & \color{red}\textbf{55.9} & \color{red}\textbf{67.6} & \color{red}\textbf{78.3} & \color{red}\textbf{86.2}  \\ \hline
Angular                 & 61.0  & 30.2  & 53.6  & 65.0  & 75.3  & 83.7  \\
Symm+Angular            & \textbf{62.3}           & \textbf{30.5}           & \textbf{54.9} & \textbf{66.9} & \textbf{77.3} & \textbf{86.0}  \\ \hline
Lifted-Struct           & 56.4           & 22.6           & 46.9  & 59.8  & 71.2  & 81.5  \\
Symm+Lifted             & \textbf{62.1}  & \textbf{28.7}  & \textbf{54.9} & \textbf{66.4} & \textbf{76.4} & \textbf{85.3}  \\ \hline
\end{tabular}%
}
\caption{Experimental results (\%) of clustering and retrieval performance on CUB200-2011 dataset in comparison with other methods. $^\dagger$ denotes the semi-hard triplet.}
\label{table:cub200}
\end{table}

\begin{table}[t!h!]
\centering
\resizebox{1.0\columnwidth}{!}{%
\begin{tabular}{c|cc|cccc}
\hline
\multirow{2}{*}{Method} & \multicolumn{2}{c|}{Clustering} & \multicolumn{4}{c}{Retrieval} \\ \cline{2-7} 
                        & NMI            & F1             & R@1   & R@2   & R@4   & R@8   \\ \hline\hline
% Contrastive             & 42.3           & 10.5           & 27.6  & 38.3  & 51.0  & 63.9  \\
% DDML                    & 41.7           & 10.9           & 32.7  & 43.9  & 56.5  & 68.8  \\ \hline
Triplet                 & 52.9           & 17.9           & 45.1  & 57.4  & 69.7  & 79.2  \\
Triplet$^\dagger$       & 55.7           & 22.4           & 53.2  & 65.4  & 74.3  & 83.6  \\
DAML (Triplet)          & 56.5           & 22.9           & 60.6  & 72.5  & 82.5  & 89.9  \\
HDML (Triplet)          & 59.4           & 27.2           & 61.0  & 72.6  & 80.7  & 88.5  \\
Symm+Triplet            & \textbf{62.4}           & \textbf{31.8}  & \textbf{69.7}  & \textbf{78.7}  & \textbf{86.1}  & \textbf{91.4}  \\
Symm+Triplet$^\dagger$  & 61.7  & 31.1           & 68.5  & 78.5  & 85.8  & 90.9  \\ \hline
N-pair                  & 62.7           & 31.8           & 68.9  & 78.9  & 85.8  & 90.9  \\
DAML (N-pair)           & 66.0           & 36.4           & 75.1  & 83.8  & 89.7  & 93.5  \\
HDML (N-pair)           & \color{red}\textbf{69.7}  & \color{red}\textbf{41.6}  & \color{red}\textbf{79.1}  & \color{red}\textbf{87.1}  & \color{red}\textbf{92.1}  & \color{red}\textbf{95.5}  \\
Symm+N-pair             & 66.3           & 36.6           & 76.5  & 84.3  & 90.4  & 94.1  \\ \hline
Angular                 & 62.4           & 31.8           & 71.3  & 80.7  & 87.0  & 91.8  \\
Symm+Angular            & \textbf{66.1}  & \textbf{35.9}  & \textbf{75.5}  & \textbf{84.0}  & \textbf{90.0}  & \textbf{94.0}  \\ \hline
Lifted-Struct           & 57.8           & 25.1           & 59.9  & 70.4  & 79.6  & 87.0  \\
Symm+Lifted             & \textbf{59.9}  & \textbf{28.5}  & \textbf{66.6}  & \textbf{77.2}  & \textbf{84.7}  & \textbf{89.9}  \\ \hline
\end{tabular}%
}
\caption{Experimental results (\%) of clustering and retrieval performance on CARS196 dataset in comparison with other methods. $^\dagger$ denotes the semi-hard triplet.}
\label{table:cars196}
\end{table}

\begin{table}[t!h!]
\centering
\resizebox{.94\columnwidth}{!}{%
\begin{tabular}{c|cc|ccc}
\hline
\multirow{2}{*}{Method} & \multicolumn{2}{c|}{Clustering} & \multicolumn{3}{c}{Retrieval} \\ \cline{2-6} 
                        & NMI            & F1             & R@1      & R@10     & R@100    \\ \hline\hline
% Contrastive             & 82.4           & 10.1           & 37.5     & 53.9     & 71.0     \\
% DDML                    & 83.4           & 10.7           & 42.1     & 57.8     & 73.7     \\ \hline
Triplet                 & 86.3           & 20.2           & 53.9     & 72.1     & 85.7     \\
Triplet$^\dagger$       & 86.7           & 22.1           & 57.8     & 75.3     & 88.1     \\
DAML (Triplet)          & 87.1           & 22.3           & 58.1     & 75.0     & 88.0     \\
HDML (Triplet)          & 87.2           & 22.5           & 58.5     & 75.5     & 88.3     \\
Symm+Triplet            & 88.9           & 30.6            & 65.7     & 81.4      & 91.7      \\
Symm+Triplet$^\dagger$  & \textbf{89.5}  & \textbf{33.9}  & \textbf{68.5} & \textbf{82.4} & \textbf{91.3} \\ \hline
N-pair                  & 87.9           & 27.1           & 66.4     & 82.9     & 92.1     \\
DAML (N-pair)           & 89.4           & 32.4           & 68.4     & 83.5     & 92.3     \\
HDML (N-pair)           & 89.3           & 32.2           & 68.7     & 83.2     & 92.4     \\
Symm+N-pair             & \color{red}\textbf{90.7}  & \color{red}\textbf{38.7}  & \color{red}\textbf{73.2} & \color{red}\textbf{86.7} & \color{red}\textbf{94.8} \\ \hline
Angular                 & 87.8           & 26.5           & 67.9     & 83.2     & 92.2     \\
Symm+Angular            & \textbf{90.5}  & \textbf{38.4}  & \textbf{73.1} & \textbf{86.6} & \textbf{94.0} \\ \hline
Lifted-Struct           & 87.2           & 25.3           & 62.6     & 80.9     & 91.2     \\
Symm+Lifted             & \textbf{90.4}  & \textbf{38.2}  & \textbf{72.3} & \textbf{86.6} & \textbf{94.2} \\ \hline
\end{tabular}%
}
\caption{Experimental results (\%) of clustering and retrieval performance on SOP dataset in comparison with other methods. $^\dagger$ denotes the semi-hard triplet.}
\label{table:sop}
\end{table}

\subsubsection{Training Speed and Memory}
% 메모리와 시간의 변화가 거의 없다. Big-O 언급 (Sampling Matters paper 참고)
The computational cost and memory consumption of symmetrical synthesis are negligible.
On a Tesla P40 GPU with a batch size of 128, forward and backward pass of training the baseline N-pair loss takes $8.852 \times 10^{-1}$ seconds, while Symm + N-pair takes $8.866 \times 10^{-1}$ seconds per batch. % 0.8852 / 0.8866
In detail, computing the baseline N-pair loss takes $0.2454$ ms, while generating symmetrical synthetic points, hard negative pair mining, and computing N-pair loss takes only $0.2497$ ms. %  0.0002454 / 0.0002497
For memory consumption of symmetrical synthesis, our proposed method requires the same size of additional matrix as the original feature points to save synthetic feature points and 16 times a similarity matrix for saving 16 possible positive and negative pairs, which are trivial.

\subsubsection{Comparison with State-of-the-Art}
% We compare our proposed method with famous metric learning losses including contrastive loss~\cite{chopra2005learning}, DDML~\cite{weinberger2009distance},
% We compare our proposed method with famous metric learning losses including triplet loss~\cite{weinberger2009distance}, triplet loss with semi-hard negative pair mining~\cite{schroff2015facenet}, N-pair loss~\cite{sohn2016improved}, angular loss~\cite{wang2017deep} and lifted structure loss~\cite{oh2016deep}, as well as hard sample generation methods, including DAML~\cite{duan2018deep} and HDML~\cite{zheng2019hardness}.
We compare our proposed method with famous metric learning losses including triplet loss, triplet loss with semi-hard negative pair mining, N-pair loss, angular loss and lifted structure loss, as well as hard sample generation methods, including DAML and HDML.
We deploy our proposed method with triplet loss, triplet loss with semi-hard negative pair mining, N-pair loss, angular loss, and lifted structure loss.
For fair comparison, we use the same pre-trained CNN model and hyper-parameters as DAML and HDML.

The experimental results on the CUB200, CARS196, and SOP datasets are listed in Tables~\ref{table:cub200}, \ref{table:cars196}, and \ref{table:sop}, respectively.
Bold numbers indicate the best score within the same type of loss, and red numbers indicate the best score within the dataset.
% Baseline loss와 비교
In comparison with the metric learning losses, combining our proposed method leads to a performance boost with high margins among all baseline losses and datasets in both clustering and retrieval tasks.
When triplet loss and lifted structure loss use Euclidean distance with $l2$-normalized features, and N-pair loss and angular loss use cosine similarity with non-$l2$-normalized features during training, the experimental results show that our proposed method is applicable to both cases.
% Hard negative generation method와 비교
Our proposed method outperforms all hard sample generation methods for every loss and dataset except one.
In the CARS196 with N-pair loss, HDML (N-pair) shows better performance than Symm + N-pair.
We speculate that this is because we use the same hyper-parameters with HDML for the fair comparison without hyper-parameter tuning.
On the other hand, the performance improvements of the existing hard sample generation methods with a large training set (i.e., SOP) are relatively smaller than with small training sets (i.e., CUB200 and CARS196), which can be critical for practical usage.
While they achieve a 2.0 to 4.6\% performance gain in Recall@1 score on the SOP dataset, our proposed method achieve a 5.2 to 11.8\% performance boost.
This demonstrates that our proposed method gives a competitive performance boost with any training set size.

\section{Conclusion}
We propose a novel method for generating synthetic hard samples, symmetrical synthesis, for deep metric learning.
Applying our method on existing metric learning losses has significantly improved performance by exploiting trivial samples with augmented information and pushing different classes away with stronger power.
We demonstrate the effectiveness of symmetrical synthesis with extensive experiments on the three famous benchmarks for image clustering and retrieval tasks.
%In the future, we will apply the idea of symmetrical synthesis to other vision tasks, such as classification, detection, and segmentation.
% Embedding space에서 기하학적 접근을 우리가 처음 했다. -> computation 적다, 직관적이다, 
% 약점을 언급해도 좋을 것 같다.

\paragraph{Acknowledgement}
% 국형근, 이태관, 박상혁, 오성준
We would like to thank Hyong-Keun Kook, Sanghyuk Park, and Tae Kwan Lee from Naver Clova vision team for helpful comments and feedback.
We are grateful to Seong Joon Oh from Naver Clova AI research team with discussions related to experiments and detailed review of the paper.

\bibliography{egbib}
\bibliographystyle{aaai}

\end{document}